\documentclass[conference]{IEEEtran}
\IEEEoverridecommandlockouts
\usepackage{cite}
\usepackage{amsmath,amssymb,amsfonts}
\usepackage{algorithmic}
\usepackage{graphicx}
\usepackage{textcomp}
\usepackage{xspace}
\usepackage{xcolor}
\usepackage{makecell}
\usepackage{booktabs}
\usepackage{multirow}
\usepackage{tikz}
\usepackage{pgfplots}
\usepackage[inline,shortlabels]{enumitem}
    \usepgfplotslibrary{groupplots}
    \pgfplotsset{compat=1.9}
    
\def\BibTeX{{\rm B\kern-.05em{\sc i\kern-.025em b}\kern-.08em
    T\kern-.1667em\lower.7ex\hbox{E}\kern-.125emX}}
\begin{document}

\title{Enhanced Fine-Tuning of Lightweight Domain-Specific Q\&A Model Based on Large Language Models\\
% \thanks{\IEEEauthorrefmark{1} Yongqian Sun is the corresponding author. }
}

% \title{Self-Evolution: A Continuous Learning Framework for Enhancing LLMs in IT Operations}

\author{
    \thanks{\IEEEauthorrefmark{1} Yongqian Sun is the corresponding author. }
    \IEEEauthorblockN{
    Shenglin Zhang\textsuperscript{1, 5}, 
    Pengtian Zhu\textsuperscript{1}, 
    Minghua Ma\textsuperscript{2},
    Jiagang Wang\textsuperscript{3}, 
    Yongqian Sun\textsuperscript{1, 6}, \\
    Dongwen Li\textsuperscript{1}, 
    Jingyu Wang\textsuperscript{1}, 
    Qianying Guo\textsuperscript{4}, 
    Xiaolei Hua\textsuperscript{4}, 
    Lin Zhu\textsuperscript{4}, 
    Dan Pei\textsuperscript{3, 7}
    } 
    
    % \IEEEauthorblockA{
    % \textsuperscript{1}\textit{Nankai University}, \textsuperscript{2}\textit{Microsoft}, 
    % \textsuperscript{3}\textit{Tsinghua University}, 
    % \textsuperscript{4}\textit{China Mobile Research Institute}, \\
    % \textsuperscript{5}\textit{Haihe Laboratory of Information Technology Application Innovation},\\
    % \textsuperscript{6}\textit{Tianjin Key Laboratory of Software Experience and Human Computer Interaction},\\
    % \textsuperscript{7}\textit{Beijing National Research Center for Information Science and Technology}
    % }

    \IEEEauthorblockA{\textsuperscript{1} \textit{Nankai University}, \{sunyongqian, zhangsl\}@nankai.edu.cn, \{zpt, lidongwen, 2320240875\}@mail.nankai.edu.cn
    }
    \IEEEauthorblockA{\textsuperscript{2} \textit{Microsoft}, minghuama@microsoft.com
    }
    \IEEEauthorblockA{\textsuperscript{3} \textit{Tsinghua University}, peidan@tsinghua.edu.cn, 13193093293@163.com
    }
    \IEEEauthorblockA{\textsuperscript{4} \textit{China Mobile Research Institute}, \{guoqianying, huaxiaolei, zhulinyj\}@chinamobile.com
    }
    \IEEEauthorblockA{\textsuperscript{5} \textit{Haihe Laboratory of Information Technology Application Innovation}
    }
    \IEEEauthorblockA{\textsuperscript{6} \textit{Tianjin Key Laboratory of Software Experience and Human Computer Interaction}
    }
    \IEEEauthorblockA{\textsuperscript{7} \textit{Beijing National Research Center for Information Science and Technology}
    }
    \vspace{-0.8cm} 
% \IEEEauthorblockA{\IEEEauthorrefmark{2}
%     \textit{Microsoft}}
% \IEEEauthorblockA{\IEEEauthorrefmark{3} 
%     \textit{Tsinghua University}}
% \IEEEauthorblockA{\IEEEauthorrefmark{4}
%     \textit{China Mobile Research Institute}}
} 
\newcommand{\method}{Self-Evolution\xspace}
\newcommand{\hq}{HQ\xspace}

\maketitle
\thispagestyle{plain}
\pagestyle{plain}

\begin{abstract}
    Large language models (LLMs) excel at general question-answering (Q\&A) but often fall short in specialized domains due to a lack of domain-specific knowledge. 
    Commercial companies face the dual challenges of privacy protection and resource constraints when involving LLMs for fine-tuning. 
    This paper propose a novel framework, \method, designed to address these issues by leveraging lightweight open-source  LLMs through multiple iterative fine-tuning rounds. 
    To enhance the efficiency of iterative fine-tuning,  \method employ a strategy that filters and reinforces the knowledge with higher value during the iterative process.
    We employed \method on Qwen1.5-7B-Chat using 4,000 documents containing rich domain knowledge from China Mobile, achieving a performance score 174\% higher on domain-specific question-answering evaluations than Qwen1.5-7B-Chat and even 22\% higher than Qwen1.5-72B-Chat.  
    %(17.58 / 6.42 - 1) = 1.74  (17.58 / 14.42 - 1) = 0.22
    \method has been deployed in China Mobile's daily operation and maintenance for 117 days, and it improves the efficiency of locating alarms, fixing problems, and finding related reports, with an average efficiency improvement of over 18.6\%.
    In addition, we release \method framework code in https://github.com/Zero-Pointer/Self-Evolution.
    
% \footnote{\IEEEauthorrefmark{1} Yongqian Sun is the corresponding author. }

% \thanks{}
% Currently, open-source large language models (LLMs) such as Qwen and LLaMA demonstrate impressive performance in general knowledge and certain specialized knowledge question-answering tasks. 
% Despite their superior performance, these models are often domain-limited.
% Instruction fine-tuning plays a crucial role in domain alignment tasks; however, acquiring instruction data is extremely costly.
% To better address these challenges, we propose the Self-Evolution framework. 
% This framework requires only a base model and knowledge documents to generate high-quality instruction data through a self-iteration process.
% During the iteration process, we use instruction-following difficulty to evaluate each instruction data, allowing us to retrieve more valuable content for further training. 
% In our experiments, the Qwen-7B model trained using Self-Evolution achieved significantly superior results, scoring 22\% higher than Qwen-72B in domain-specific question-answering evaluations.
% We subsequently deployed this framework on China Mobile's internal knowledge base, where the model's autonomous learning yielded practical outcomes consistent with our experimental findings.
\end{abstract}


\section{Introduction}
With the emergence of large language models (LLMs) such as Qwen~\cite{bai2023qwen}, LLaMA~\cite{touvron2023llama}, and GPT~\cite{brown2020language}, their exceptional generation, understanding of complex language structures and dialogue capabilities have garnered widespread attention~\cite{devlin2018bert, raffel2020exploring}.
However, in specific domains, their performance often fails to meet practical requirements. 
For instance, GPT-4 may cite incorrect legal provisions when answering legal questions, leading to erroneous analytical conclusions.
ChatLaw-MoE~\cite{cui2023chatlaw}, fine-tuned on high-quality law data, has outperformed GPT-4 across multiple application scenarios. 
Therefore, enabling general models to acquire domain-specific knowledge allows for deploying a domain model with minimal computational resources, potentially outperforming general models with ten times the number of parameters.
% With instruction alignment~\cite{brown2020language}, their conversational patterns closely mirror human dialogue styles. 
% The impact of these models extends beyond the academic domain of Natural Language Processing (NLP), exerting profound influence in industrial sectors. 
% They have made significant strides in advancing human-computer interaction, intelligent customer service, and virtual assistant technologies, demonstrating immense potential in these areas.

State-of-the-art approaches extensively utilize instruction fine-tuning (IFT) to align general-purpose models with specific application domains and maximize their effectiveness. 
InstructGPT~\cite{ouyang2022training} employed instruction fine-tuning to bridge the performance gap between models with a hundredfold difference in parameter count.
In the absence of instruction data, certain approaches~\cite{taori2023stanford, peng2023instruction, vicuna2023, xu2023baize}  use advanced LLMs to construct instruction datasets, achieving performance close to GPT-3.5 and GPT-4. 
However, these methods cannot guarantee the correctness and diversity of the generated instruction data.
Fortunately, high-quality instruction data is scarce in most scenarios, while the volume of knowledge documents is enormous.

% Self-alignment techniques~\cite{wang2022self, zhang2023self, sun2024principle} have liberated instruction data assembly from external LLMs, capitalizing on the extensive domain knowledge base.

In summary, applying general-purpose models to specific domains presents the following challenges:
\begin{enumerate}
    \item \textbf{Limitation of Computational Resources.}
    Model performance is typically proportional to the scale of the model's parameters. 
    However, fine-tuning and deploying powerful general-purpose language models requires substantial computational resources. 
    For example, a LLM with 72B parameters using fp16 precision requires five Tesla V100-32GB GPUs for inference. 
    Fine-tuning such a model incurs even greater costs. 
    This is prohibitively expensive and impractical for tasks that must be continuously available.
    \item \textbf{High-quality data scarcity.}
    Domain-specific high-quality instruction data is often scarce.
    Manually correcting instruction data requires significant human effort, making it expensive.
    A solution is needed to automatically construct high-quality data without human assistance.
    \item \textbf{Lack of diversity and correctness.}
    Firstly, using a fixed model to construct instruction data tends to generate overly similar data. 
    Additionally, relying solely on the model's internal capabilities for data generation may result in incorrect or irrelevant data for the domain. 
    The model might need more domain understanding or have learned incorrect knowledge, leading to hallucination issues.
    We hope the model can dynamically learn from unsupervised domain documents, continually improving its capabilities while ensuring the diversity and accuracy of data generation.
    \item \textbf{Data privacy.}     
    Due to the inclusion of private information in domain-specific data, fine-tuning commercial LLMs poses major challenges when dealing with sensitive internal company data, including privacy leakage and high costs.

\end{enumerate}
In this paper, we propose a novel framework \textbf{\method} to address the aforementioned challenges.
The contributions of this paper are summarized as follows:
\begin{enumerate}
    \item 
   Considering the costs and privacy concerns during actual deployment, we select an open-source model with 7B parameters as the data generation, scoring model, and model for QA tasks in real scenarios. All phases in \method can be completed with just one Tesla V100-32GB GPU, significantly reducing computational resource requirements. (Addressed challenges 1 and 4.)
   \item \method uses LLM to generate instrution data based on a large number of unlabeled knowledge documents, ensuring domain relevance and correctness while avoiding the need for manual assistance. 
   Additionally, the LLM undergoes iterative updates, generating a new batch of data each time. 
   This process ensures diversity between different batches of data. (Addressed challenges 2 and 3.)
   \item We conducted extensive evaluation experiments using real-world data from China Mobile, a top-tier telecommunications provider providing services for one billion+ monthly active users (MAU). 
   \method achieves a performance score 174\% higher on domain-specific question-answering evaluations than without using \method and even 22\% higher than Qwen1.5-72B-Chat.  %(17.58 / 6.42 - 1) = 1.74  (17.58 / 14.42 - 1) = 0.22
   The \method has been deployed in China Mobile's daily operation and maintenance for 117 days. %(3.4 -> 6.28 = 117 days)

\end{enumerate}

\section{Related Work}
\subsection{Instruction Fine-tuning}
The potential of LLMs in the specific domain is vast and promising. 
For example, Microsoft deployed GPT to summarize anomalous events in its services~\cite{jin2023assess}.
However, as task complexity and requirements increase, instruction fine-tuning (IFT) is widely adopted to enhance model performance.
FLAN~\cite{wei2021finetuned} achieved significant improvements in generalization by fine-tuning a high-quality instruction dataset. 
InstructGPT~\cite{ouyang2022training} successfully aligned GPT-3~\cite{brown2020language} with human intent by fine-tuning a dataset rich in real-world instruction forms and task types. 
OWL~\cite{guo2024owl} collected numerous operation domain instructions and achieved remarkable results in log parsing and anomaly detection.
However, these methods require a large amount of manually annotated data, which becomes a bottleneck for widespread application due to the high cost.

\subsection{Instruction Data Generation}
Researchers have extensively explored methods to reduce human involvement in generating instruction data. Some methods~\cite{taori2023stanford,xu2023wizardlm,vicuna2023,peng2023instruction} use advanced commercial models to create instruction datasets. 
For instance, Alpaca~\cite{taori2023stanford} uses a small amount of manually constructed data to extract knowledge from DaVinci-003~\cite{hinton2015distilling}, creating a 52k instruction dataset. 
It fine-tunes LLaMA to achieve performance close to GPT-3.5. 
Peng et al.~\cite{peng2023instruction} extract knowledge from GPT-4, resulting in higher quality and more diverse responses.

Another class of methods~\cite{wang2022self,sun2024principle,xu2023baize} employs a self-guided approach. 
These methods extract knowledge from the model and then use this newly constructed data to enhance domain or task capabilities. 
Self-Instruct~\cite{wang2022self}, for instance, proposes using self-generated samples to enhance the instruction-following ability of pre-trained language models. Self-Align~\cite{sun2024principle} mainly adopts topic-guided red-blue adversarial self-guidance and principle-driven self-calibration to construct data and fine-tune models, requiring less than 300 lines of manually constructed data (including 195 seed prompts, 16 principles, and five examples) to achieve high-quality fine-tuned model.
The potential of these self-guided methods is certainly worth exploring further.

However, these methods still require manually constructed supervision data and are limited by the model's inherent knowledge constraints, preventing them from generating instruction data beyond the model's capabilities.

\subsection{Instruction Data Selection}
In the early stages of IFT research, many works improved model capabilities by building large instruction datasets. 
However, LIMA~\cite{zhou2024lima} proposed that ``less alignment is more" showing that fine-tuning the model with only 1,000 high-quality samples can achieve a performance comparable to GPT-4. 
Appropriate data filtering strategies can improve learning efficiency and help reduce hallucinations caused by overtraining~\cite{touvron2023llama}.
%~/cite{huang2023survey}

% Instruction mining~\cite{cao2023instruction} uses statistical regression for data selection. 
% ALPAGASUS~\cite{chen2023alpagasus} uses ChatGPT for scoring, though this method may overlook the target model's capabilities and lacks interpretability.
% The forgetting score~\cite{toneva2018empirical} tracks changes in sample classification during training.
% GraNd~\cite{paul2021deep} prunes data based on the gradient norm of the sample. 
% While the forgetting score and GraNd require significant overhead since they need to continuously update the scoring model, thus increasing the overall model training time.

ALPAGASUS~\cite{chen2023alpagasus} uses ChatGPT for scoring but might miss the target model's strengths and lacks clarity.
The forgetting score~\cite{toneva2018empirical} monitors shifts in sample classification during training.
GraNd~\cite{paul2021deep} trims data based on the sample's gradient magnitude.
Both forgetting score and GraNd are costly, as they need constant model updates, prolonging training time.

Instruction Following Difficulty (IFD)~\cite{li2023quantity} stands out for its efficiency, using the representation features of the target model to identify high-quality instruction data.
It provides a simpler, cheaper, and interpretable approach by computing the generation complexity of the answer using a single fixed scoring model.

\begin{figure}[!htb]
\centerline{\includegraphics[width=1\linewidth, keepaspectratio]{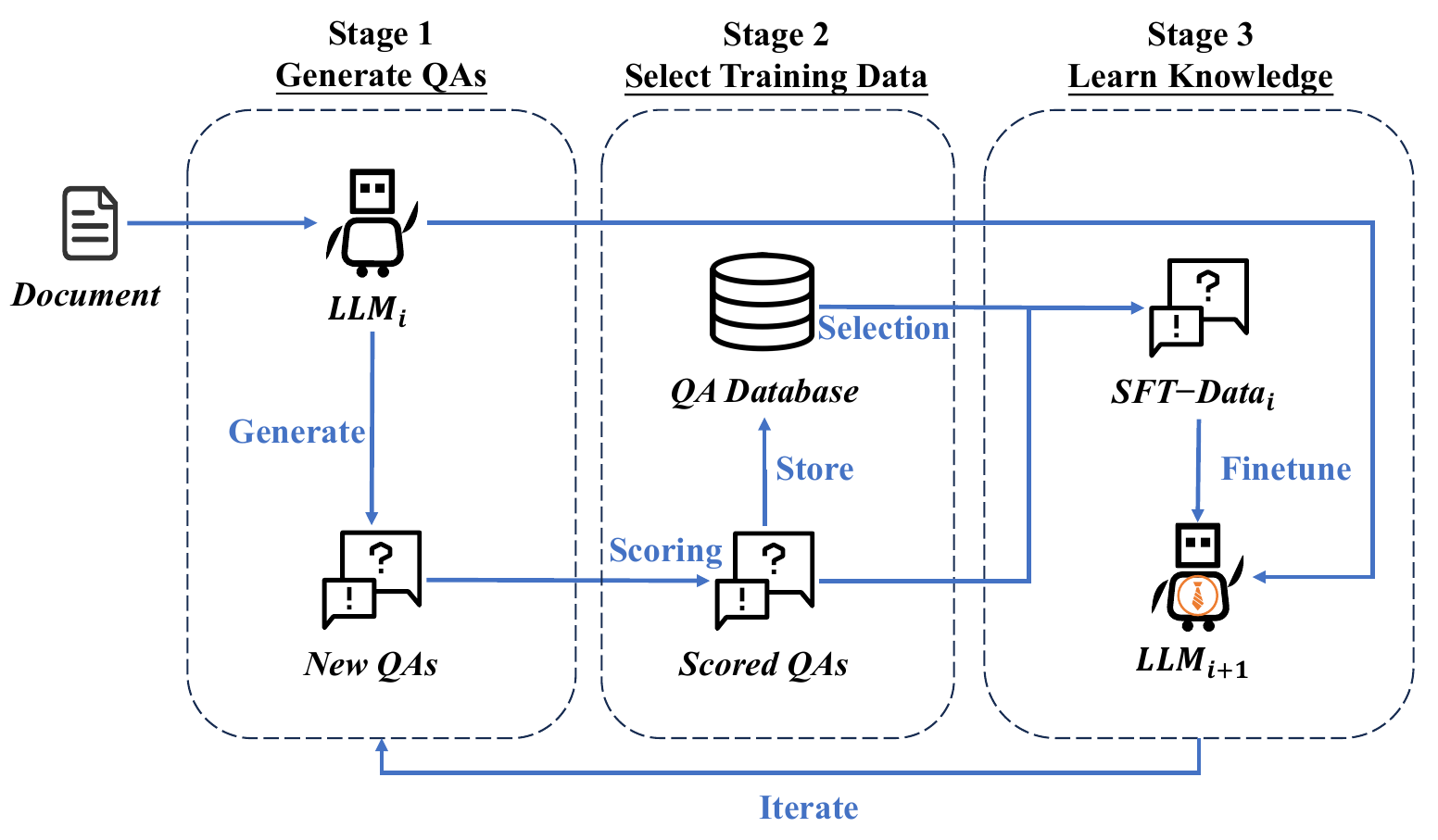}}
\caption{Self-Evolution}\label{fig-se}
\end{figure}

\section{Method}

The overview of \method is illustrated in \figurename~\ref{fig-se}. 
To start, \method requires a LLM $\theta_0$ as the initial QA\footnote{As the instruction data in this paper consistently takes the form of question-answer pairs, the terms ``instruction data", ``QA" and ``question-answer pairs" are used interchangeably in the following text.} generation model and scoring model and a collection of domain-related documents $T$.
\method consists of  three phases.
In the first stage, the QA generation model generates QA pairs based on the domain-related documents.
In the second phase, the scoring model and a scoring metric are employed to identify valuable samples from all historical instruction QA pairs.
In the third phase, these valuable instruction samples are used to conduct a new round of IFT, reinforcing the model's domain knowledge.
These three phases iterate continuously until the desired performance is achieved.
The following sections will provide a detailed description of these phase.
% \method is illustrated in \figurename~\ref{fig-se}. 
% The initialization of Self-Evolution requires only an initial model $\theta_0$ and a collection of documents $T$.
% In the first phase, the model is tasked with generating corresponding QA pairs from knowledge documents.
% The second phase begins by filtering the historical instruction data to identify valuable samples for continued learning.
% The third phase combines instruction data from the first and second phases to conduct the current round of IFT, effectively consolidating knowledge within the model.
% These three phases will iterate continuously until the model achieves the desired performance.
% The following sections will provide a detailed explanation of these three processes.

\subsection{QA generation}\label{sec-qa-gene}
More new QA data are generated in the QA generation phase. 
\method constructs new questions and answers based on each domain-related document rather than deriving them from manually constructed questions. 

This questions generation process is represented as $q_{ij} = LLM(\theta_i, t_j)$, where $t_j$ is the j-th document in $T$.
During this process, we design delicated prompt to prioritize two key aspects:
\begin{enumerate*}
    \item Question conciseness: Preventing the generation of content with multiple sub-questions, which could lead to model hallucinations (Note 2).
    \item Question validity: Ensuring each generated question is answerable (Note 6).
\end{enumerate*}
The detailed prompt used for question generation is as follows:
\begin{table}[!htb]
\centering
\caption{Question generation prompt.}
\label{tab-q-prompt}
\resizebox{1.\columnwidth}{!}
{
\begin{tabular}{|l|} 
\hline
\textbf{\textcolor{blue}{Domain Knowledge:}}\\
Reference document: \{Knowledge\} \\
\textbf{\textcolor{blue}{Role Description:}}\\
You are an expert in the operations domain.  \\
Based on your comprehensive knowledge and the information provided above......  \\
\textbf{\textcolor{blue}{Rules Description:}}\\
Note 1: The question should be as concise as possible. \\
Note 2: The question should not contain multiple sub-questions, only one question is permitted. \\
...... \\
\textbf{Note 6: Do not output declarative sentences; it must be a question!} \\
Please formulate a question now. \\
Question: \\
\hline
\end{tabular}}
\end{table}

This answer generation process is represented as $a_{ij} = LLM(\theta_i, t_j, q_{ij})$. 
Incorporating $t_j$, ensures that the questions are correctly answered.
In this process, we emphasize response completeness, ensuring that the generated content is a complete answer rather than one containing pronouns referring back to the document. %~\cite{sun2024principle}
The prompt used for answer generation is as follows:
\begin{table}[!htb]
\centering
\caption{Answer generation prompt.}
\label{tab-a-prompt}
\resizebox{1.\columnwidth}{!}
{
\begin{tabular}{|l|} 
\hline
\textbf{\textcolor{blue}{Role Description:}}\\
You are an expert in the field of operations...... \\
You must generate responses based on the requirements. \\
\textbf{\textcolor{blue}{Workflow Description:}}\\
1. Receive and parse the user's question. \\
2. Read and analyze the document provided by the user. \\
3. Provide a concise and comprehensive answer by combining your knowledge  \\
with the document content. \\
\textbf{\textcolor{blue}{In Context Learning:}}\\
Examples: \\
Question: Which is the largest planet in the solar system? \\
Knowledge fragment: The solar system consists of eight planets, with Jupiter  \\
being the largest. Its mass is 2.5 times that of all other planets combined. \\
Answer: The largest planet in the solar system is Jupiter. \\
\textbf{\textcolor{blue}{Warnings:}}\\
Your answer will be sent independently of the document after generation......  \\
Your response must ensure two points: conciseness and accuracy. \\
\textbf{\textcolor{blue}{Domain Knowledge and Question:}}\\
Question: \{Question\} \\
Knowledge fragment: \{Knowledge\} \\
\hline
\end{tabular}}
\end{table}

% You are an expert in the field of operations, responsible for answering various \\
% questions in the communications domain.  \\
% You must generate responses based on the requirements. \\
% Workflow: \\
% 1. Receive and parse the user's question. \\
% 2. Read and analyze the document provided by the user. \\
% 3. Provide a concise and comprehensive answer by combining your knowledge  \\
% with the document content. \\
% Examples: \\
% Question: Which is the largest planet in the solar system? \\
% Knowledge fragment: The solar system consists of eight planets, with Jupiter  \\
% being the largest. Its mass is 2.5 times that of all other planets combined. \\
% Answer: The largest planet in the solar system is Jupiter,  \\
% with a mass 2.5 times that of all other planets combined. \\
% Warnings: \\
% Your answer will be sent independently of the document after generation,  \\
% so do not use the word "document" in your response.This would confuse users  \\
% as they cannot see the corresponding document and can only read your answer. \\

% Your response must ensure two points: conciseness and accuracy. \\
% Question: \{Question\} \\
% Knowledge fragment: \{Knowledge\} \\

After obtaining the newly generated questions and answers, \method combines them into new instruction data $D_i = \{(q_{i0}, a_{i0}), (q_{i1}, a_{i1}), \ldots, (q_{i|T|}, a_{i|T|})\}$.
% The two-stage QA pair generation allows for the creation of high-quality, context-relevant QA pairs that serve as robust training data for the subsequent IFT process.

\subsection{Data Selection And Training}
Prior to conducting the i-th round of IFT, we can filter and select a subset of instruction data from the previous $i-1$ rounds to enhance the training process.
\method employs the IFD metric~\cite{li2023quantity} to identify more valuable instruction data.
Equation \ref{eq-ifd} represents the calculation method for the IFD score, while Equations \ref{eq-ifd-qa} and \ref{eq-ifd-a} denote the Conditioned Answer Score and Direct Answer Score, respectively.
\begin{gather}
s_\theta(A \mid Q)=-\frac{1}{N} \sum_{i=1}^N \log P\left(w_i^A \mid Q, w_1^A, \ldots, w_{i-1}^A ; \theta\right) \label{eq-ifd-qa} \\
s_\theta(A)=-\frac{1}{N} \sum_{i=1}^N \log P\left(w_i^A \mid w_1^A, w_2^A, \ldots, w_{i-1}^A ; \theta\right) \label{eq-ifd-a} \\
\operatorname{IFD}_\theta(Q, A)=\frac{s_\theta(A \mid Q)}{s_\theta(A)} \label{eq-ifd}
\end{gather}

The Conditioned Answer Score quantifies a model's ability to produce responses that align with both the given instructions and the correct answers. 
It assesses the model's output congruence with the directive and the expected solution.
The Direct Answer Score evaluates the LLM's capacity to independently generate correct answers, reflecting the answer's intrinsic complexity in the absence of contextual instructions.
A high IFD score indicates the model's difficulty in aligning responses with instructions, thereby highlighting the instruction's complexity.

Therefore, \method extract $k$ instruction data with the highest IFD scores from $D_0, D_1, \ldots, D_{i-1}$ to form $IFD_i$. 
This set $IFD_i$ is then combined with $D_i$ for the i-th round of training, leveraging historical high-quality data alongside newly generated data, potentially enhancing the efficiency and effectiveness of each training iteration.

% After obtaining the data required for the i-th round of training, we fine-tune $\theta_{i}$ and proceed to the next iteration.

\subsection{Next Iteration}
\method employs a model self-evolution scheme.
To elucidate the principles underlying this scheme, we define a scoring function $score = f(q, a)$ that evaluates the quality of an answer $a$ with respect to a question $q$.
As previously mentioned, $a = LLM(\theta_i, q)$ denote the response of model $\theta_i$ to $q$, and $a = LLM(\theta_i, t, q)$ represent the response of model $\theta_i$ to $q$ given a highly relevant knowledge document $t$.
We define $\theta_{i+1} = IFT(\theta_i, q, a)$ as the next-generation model $\theta_{i+1}$ resulting from fine-tuning $\theta_i$ on the instruction data pair $(q, a)$.
We leverage In-context Learning~\cite{liu2023pre} to establish the first inequality:
\begin{equation}
\label{eq-score1}
f(q, LLM(\theta_i, q)) \leq f(q, LLM(\theta_i, t, q))
\end{equation}
This inequality demonstrates that the instruction data $((q, LLM(\theta_i, t, q))$ provides valuable learning opportunities for model $\theta_i$.
Consequently, we derive $\theta_{i+1}$ through $\theta_{i+1} = IFT(\theta_i, q, a)$. 
Post-training, we obtain the second inequality:
\begin{equation}
\label{eq-score2}
f(q, LLM(\theta_{i}, q)) \leq f(q, LLM(\theta_{i+1}, q))
\end{equation}
Thus, model $\theta_i$ completes one iteration of evolution.
The iterative process can be terminated by setting an iteration threshold. 
Empirically, this threshold is proportionally related to the model's parameter count and inversely related to the data volume. 
Smaller parameter counts tend to be more susceptible to hallucinations, necessitating threshold adjustments based on both parameter count and data volume.% ~\cite{gekhman2024does}

\section{Experimental Setup}
\subsection{Model and Dataset}
The base model selected for our experiments is Qwen1.5-7B-Chat~\cite{bai2023qwen}, denoted as $\theta_{0}$.
We use the LoRA (Low-Rank Adaptation)~\cite{hu2021lora} method to fine-tune models.
The LoRA hyperparameters are configured as follows: lora-rank is set to 4, and lora-alpha is set to 8. 
Notably, we set lora-target to ``all"~\cite{xu2023baize}, which enables us to achieve superior training results.
The model chosen for IFD scoring is Qwen1.5-7B-Chat, denoted as $\theta_{ifd}$. 
It is important to note that $\theta_{ifd}$ does not participate in the subsequent training process.
Its parameters remain fixed throughout the iteration process, ensuring consistent scoring criteria in each round of evaluation.

We select 4,000 valuable internal knowledge documents from China Mobile, denoted as $T$, where $|T| = 4000$.
As shown in Table~\ref{tab-docu-exp}, $T$ contains crucial operational knowledge such as alert analysis, configuration analysis, and operational experience, enabling operation engineers to quickly familiarize themselves with and solve problems.
These knowledge documents are incorporated into the training process. 
Specifically, they are converted into corresponding instruction data and subsequently used for IFT.
\begin{table}[!htb]
\centering
\caption{Example for Knowledge Document.}\label{tab-docu-exp}
\resizebox{1.\columnwidth}{!}
{
\begin{tabular}{l} 
\hline
Alarm: \{Alarm instance\} \\
Alarm Explanation: \\
-\{This is the reason for the alarm to appear\} \\
-\{This is the condition for the alarm to be cleared\} \\
-\{This is the specific threshold for the occurrence and resolution of alarms\} \\
Possible Reasons: \\
-Reason 1: \{This is the first possible reason that may occur\} \\
-Reason 2: \{This is the second possible reason that may occur\} \\
Processing Steps: \\
-Reason 1: \\
\hspace{4mm}-\{Step 1\}, \{Step 2\}... \\
% \hspace{4mm}-\{Step 2\} \\
-Reason 2: \\
\hspace{4mm}-\{Step 1\}, \{Step 2\}... \\
\hline
\end{tabular}}
\end{table}

We collected 100 real-world question-answer pairs related to these documents, as shown in Table~\ref{tab-qa-exp}. 
These pairs correspond to the knowledge that operations engineers need to acquire on an ad hoc basis during their work. 
Due to their close association with the knowledge contained in the documents, we consider this set as a test set to evaluate the model's performance.
\begin{table}[!htb]
\centering
\caption{Example for Question and Answer.}\label{tab-qa-exp}
\resizebox{1.\columnwidth}{!}
{
\begin{tabular}{l} 
\hline
\textbf{Question:} \\
How to gradually troubleshoot and solve the problem when device A starts \\
and device B cannot function properly? \\
\hline
\textbf{Answer:} \\
When the alarm of B not working properly appears after device A is started, \\
the following steps can be followed for troubleshooting and handling: \\
1. Check component C. \\
\hspace{4mm}-If C is firm, proceed to step 3. \\
\hspace{4mm}-If it is not secure, try reinstalling component C. \\
2. After reinstalling component C, check if the alarm disappears. \\
\hspace{4mm}-If the alarm disappears, the problem has been resolved, and the process ends. \\
\hspace{4mm}-If the alarm still exists, proceed to step 3. \\
3. Check if component C is damaged. \\
\hspace{4mm}-If damaged, proceed to step 4. \\
\hspace{4mm}-If not damaged, please contact technical personnel. \\
4. Replace component C with a new one and check if the alarm is cleared. \\
Throughout the entire process, it is essential to backup data before \\
operation to prevent data loss. \\
\hline
\end{tabular}}
\end{table}

\subsection{Baseline}

\begin{table}
    \centering
    \caption{Comparison of different baseline methods.}\label{tab-baseline}
    \resizebox{1.\columnwidth}{!}{
    \begin{tabular}{c | cc}
        \hline
        Model Name & Is Aligned? & Data Source\\
        \hline
        Qwen1.5-7B-HQ & Yes & \makecell[c]{Generated by Qwen1.5-72B-Chat \\ with documents} \\
        \hline
        Qwen1.5-7B-Chat & No & - \\
        Qwen1.5-72B-Chat & No & - \\
        GPT-3.5 & No & - \\
        \hline
    \end{tabular}}
\end{table}

\subsubsection{Qwen1.5-7B-Chat-Fine-Tuned by High Quality QA}
The Qwen1.5 series of language models has demonstrated exceptional performance in the Chinese language domain ~\cite{bai2023qwen}, with Qwen1.5-72B-Chat achieving capabilities comparable to GPT-3.5 on certain tasks.
Consequently, we utilized Qwen1.5-72B-Chat to generate 4,000 high-quality question-answer pairs following the approach outlined in Section~\ref{sec-qa-gene}. 
These pairs were subsequently used to train a Qwen1.5-7B-HQ model for evaluation purposes, denoted as $\theta_{HQ}$.
This methodology of extracting knowledge from documents using a superior model emulates the industrial scenario of constructing IFT data from operational documentation, which often yields favorable results~\cite{taori2023stanford, vicuna2023}.
\subsubsection{Original LLM}
We employed the untrained Qwen1.5-7B-Chat and Qwen1.5-72B-Chat models in our evaluation to simulate the scenario of using open-source models directly for domain-specific question answering.
Additionally, we included GPT-3.5 in our evaluation to simulate the scenario of utilizing a closed-source model for domain-specific question answering.

\section{Evaluation Metrics}
We use the BLEU~\cite{papineni2002bleu} score of $\theta_{HQ}$, denoted as $BLEU(\theta_{HQ})$, as a benchmark score and calculate the relative scores of other models in comparison to it.
The performance score for a model $\theta$ is calculated as:
\begin{equation}\label{eq-bleu}
Score = \frac{BLEU(\theta)}{BLEU(\theta_{HQ})}
\end{equation}

To better illustrate the differences between methods, we let $\theta_{HQ}$ serves as a target model for comparison.
We collected 100 valuable subjective questions internally from China Mobile, which are related to the knowledge documents $T$. 
These questions can reflect the model's learning of $T$ through question-answering performance.
This score represents how closely a given model's performance in the domain-specific task approaches that of the optimally fine-tuned model $\theta_{HQ}$.

\section{Experimental Results}

\begin{figure}[!htb]
\vspace{-0.4cm} 
\centerline{\includegraphics[width=\linewidth, keepaspectratio]{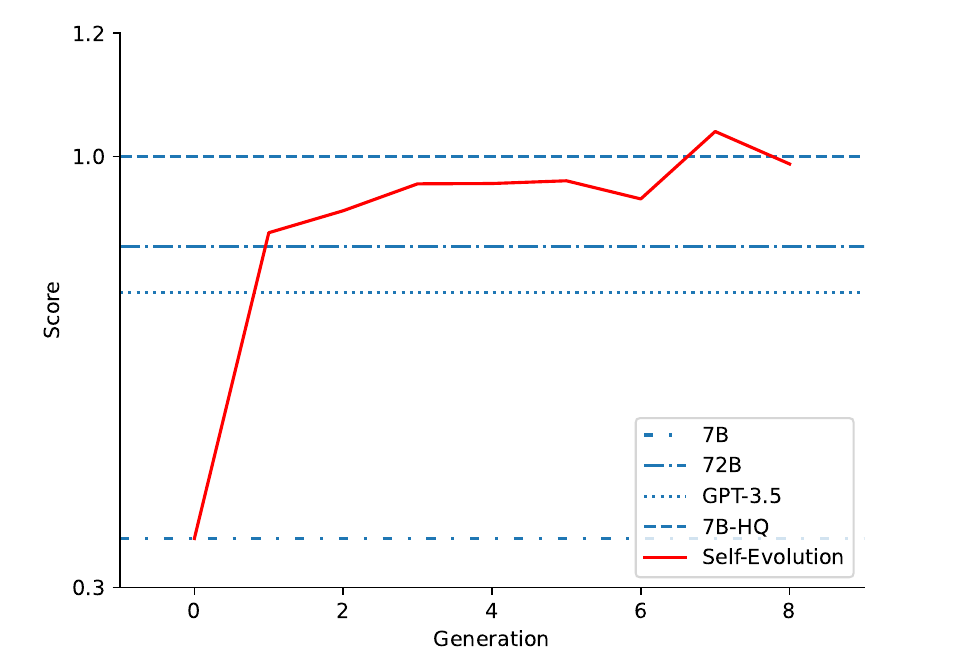}}
\caption{The x-axis represents the number of iterations of the \method, while the y-axis shows the performance scores of different models. The horizontal lines in the graph represent the performance of four distinct models, and the line graph depicts the performance of the \method at each iteration.}\label{fig-exp-res}
\end{figure}
We compare Qwen1.5-7B-Chat, trained using \method, with multiple baseline models.
As shown in Figure~\ref{fig-exp-res}, untrained models perform poorly in domain-specific knowledge question answering tasks.
The $\theta_{HQ}$ model, fine-tuned with high-quality data, demonstrates excellent performance.
Notably, \method surpasses the performance of both GPT-3.5 and Qwen1.5-72B-Chat in its first iteration.
As the iterations progress, the model's performance gradually approaches that of $\theta_{HQ}$, ultimately surpassing it by the seventh round.
Based on the above experiments, we can conclude that \method enables Qwen1.5-7B-Chat to surpass the performance of Qwen1.5-72B-Chat-assisted alignment. 
This demonstrates the effectiveness of the proposed method.

\section{Ablation Experiment}
\vspace{-0.4cm} 
\begin{figure}[!htb]
\centerline{\includegraphics[width=\linewidth, keepaspectratio]{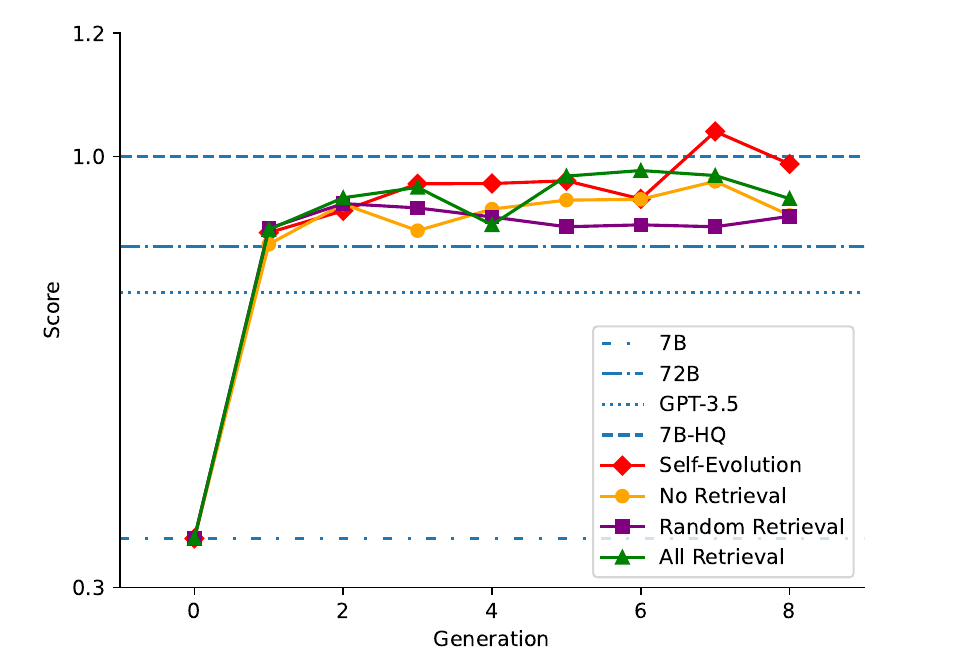}}
\caption{Ablation Experiment Results. }\label{fig-abl-allinone}
\end{figure}

\subsection{Historical Data Retrieval Module}
One of the core components of \method is the historical data retrieval module. 
To investigate its specific role, we designed targeted experiments.
After generating the instruction data $D_i$ in the i-th iteration, instead of performing historical data retrieval, we directly used it as the complete training dataset.
The results, as shown in Figure \ref{fig-abl-allinone}, indicate that the iterated model failed to surpass the performance of \hq, and the training effectiveness was compromised to some extent.
This demonstrates that historical instruction data is valuable and needs to be retrieved and relearned.

\subsection{Historical Data Retrieval Strategy}
To validate the effectiveness of using IFD scores for efficient historical instruction data filtering in \method, we designed two experiments.

To demonstrate our data filtering's logic, the corresponding experiment used all previously generated data for training.
The results, as shown in Figure \ref{fig-abl-allinone}, indicate that the performance of the iterated model rapidly deteriorated.
Training with too much data caused the model to hallucinate. 
In the eight-iteration experiment, the full retrieval strategy took about three times longer than \method.
This proves that discarding a portion of the data not only accelerates training speed but also enhances training effectiveness.
% \begin{figure}[!htb]
% \centerline{\includegraphics[width=\linewidth, keepaspectratio]{imgs/All Retrieval_eval_perc.pdf}}
% \caption{Full Retrieval of Historical Instruction Data}\label{fig-abl-all}
% \end{figure}

To demonstrate the superiority of our data filtering strategy, we designed an experiment using a random retrieval strategy during the recall phase, where $k$ instruction data were randomly recalled from historical instruction data and added to the training set.
Figure \ref{fig-abl-allinone} shows performance gains only in the first two generations, with further training harming results.
This indicates the need for a proper data filtering strategy, as an unstable retrieval approach can degrade model performance.

% \begin{figure}[!htb]
% \centerline{\includegraphics[width=\linewidth, keepaspectratio]{imgs/Random Retrieval_eval_perc.pdf}}
% \caption{Random Retrieval of Historical Instruction Data}\label{fig-abl-random}
% \end{figure}

In the aforementioned experiments, we tested three alternative approaches: removing the historical data retrieval module, employing a full retrieval strategy, and using a random retrieval strategy. 
All of these approaches resulted in some degree of performance degradation compared to \method.
These results demonstrate that the data retrieval module in \method is essential, and the data filtering strategy centered on IFD plays a crucial role in the method's effectiveness.

\section{Conclusion and Future Work}
In this paper, we address a key challenge in applying LLMs to the specific domain: the difficulty in utilizing vast amounts of unlabeled knowledge documents.
To tackle this issue, we employ self-alignment in \method to rapidly construct a large volume of instruction data. 
As the iteration progresses, both the model's capabilities and the quality of generated data improve.
To maximize the utilization of instruction data generated in each iteration, we use IFD scores to filter out high-quality data to assist in training.
In the China Mobile business question-answering evaluation, our approach, using only a 7B model throughout, outperforms solutions assisted by 72B models, conserves a significant amount of computational resources.

In current business scenarios, multi-turn dialogue capabilities are becoming increasingly important. 
Therefore, in future work, we plan to extend \method to improve the model's domain-specific multi-turn dialogue capabilities using only unsupervised text data.

\section*{ACKNOWLEDGMENTS}
This work is supported by the National Natural Science Foundation of
China (62272249, 62302244, 62072264).

\vspace{12pt}
\bibliographystyle{IEEEtran}
\bibliography{reference.bib}

\end{document}